# Omnifont Persian OCR System Using Primitives

Azarakhsh Keipour[*], Mohammad Eshghi[**],
Sina Mohammadzadeh Ghadikolaei[***]
Faculty of Electrical and Computer Engineering
Shahid Beheshti University
Tehran, Iran
[*]keipour@ieee.org, [**]m-eshghi@sbu.ac.ir,
[***]mohammadzadeh.sina@gmail.com

Negin Mohammadi
Computer Engineering Department
Sharif University of Technology
Tehran, Iran
negin.mhdi@gmail.com

Shahab Ensafi
School of Computing
National University of Singapore
Kent Ridge, Singapore
shahab.ensafi@gmail.com

*Abstract*—**In this paper we introduce a model-based omnifont Persian OCR system. The system uses a set of 8 primitive elements as structural features for recognition. First, the scanned document is preprocessed. After normalizing the preprocessed image, text rows and sub-words are separated and then thinned. After recognition of dots in sub-words, strokes are extracted and primitive elements of each sub-word are recognized using the strokes. Finally, the primitives are compared with predefined set of character identification vectors in order to identify sub-word characters. The separation and recognition steps of the system are concurrent, eliminating unavoidable errors of independent separation of letters. The system has been tested on documents with 14 standard Persian fonts in 6 sizes. The achieved precision is 97.06%.**

I. INTRODUCTION

Maintaining paper documents is expensive and space consuming. To reduce costs of keeping these documents, we can convert them into electronic documents. On the other hand, simply keeping scanned documents in image format, consumes a lot of permanent memory, but still the resulting files are not searchable. As a result, there is a growing need to keep documents in text format to reduce memory storage requirements and to add searching capability. There are two possible solutions to this problem: manually re-typing the paper documents, or directly converting them into text. The latter is called Optical Character Recognition (OCR).

Nowadays, OCR is widely used in industry. Its applications range from barcode recognition and reading serial numbers to reading destination of packages.

The first research on OCR for Persian language is reported in 1975 [1] and the first algorithm is proposed in 1978 [2]. The first OCR algorithms for Arabic language are proposed in 1980 [3, 4]. The main difference between Persian and Arabic text is the addition of four characters "پ" ,"چ" ,"گ" and "ژ" in Persian alphabet. The similarity of the alphabets makes the proposed OCR algorithms applicable to both languages, thus we do not differentiate between the alphabets of these languages in this paper.

There are two main approaches to recognition of Persian and Arabic characters. The first is the use of statistical features, such as the number of pixels in vertical and horizontal projections. Another approach is the use of structural features, such as the shape of characters or the number of closed curves in a character. This approach is generally more robust to font changes.

Since 1980, many OCR algorithms for Persian and Arabic text recognition have been proposed. Most of them use statistical features. Among the structural methods, some use a set of primitive elements that can form all Persian and Arabic letters. In [5] a set of 10 primitives was used with neocognitron type of neural network classifier, which was tested successfully for handwriting recognition. In [6] a set of 9 primitives was used for recognition of printed sub-words with at most three letters. In [7] the same set of primitives with a slightly different approach was used to recognize printed sub-words with any number of letters. The sets of primitives in all of the proposed methods are similar, and the differences are mostly in their approaches. The proposed OCR system uses a set of 8 primitives to recognize Persian and Arabic characters in three main steps. It has very little dependency on the size and type of the fonts. The proposed system accomplishes two processes of separation and recognition simultaneously, which eliminates errors occurring in complete separation before the recognition phase.

In section II, we discuss the characteristics of Persian/Arabic text. In section III, our OCR system is introduced. In section IV we discuss the test results of the proposed OCR system. Finally, in section V we conclude with a discussion on the obtained results.

II. BACKGROUND

Some of the characteristics of Persian/Arabic text make character recognition of these languages much harder than Latin languages (e.g. English). Since our OCR system is specialized mostly for Persian text, from now on we will concentrate on Persian character recognition. Although, modifying the system for recognition of Arabic text is an easy task.

Some of the main characteristics of Persian text are as follows:

- Persian text is written from right to left.

- Each word in Persian consists of one or more sub-words.

- There are 32 letters in Persian alphabet. Each of the letters can have up to 4 forms, according to their position in sub-word. All of the 32 letters have *separate* and *end* forms. 25 of them also have *first* and *middle* forms. Thus, the total number of different letter forms in Persian alphabet is 114, which makes recognition even harder than it seems. Furthermore, there are some signs adopted from Arabic language, which are common in Persian text, including "harkat" signs ( ِ َ ُ ), "tanvin" signs ( ٍ ً ٌ ), "tashdid" sign ( ّ ), "mad" sign ( ~ ) and "hamzeh" ( ٔ ء ). Table I shows all of the common letter forms and signs in formal Persian text.

- Persian text is written above a horizontal line called *baseline*. Fig. 1 shows a sample Persian sentence with its baseline.

- Most of the Persian letters have dots in their forms. In fact, 50 of 114 letter forms don't have any dots, 36 of them have only one dot, 10 of them have two dots and 18 of them have three dots. Many of the letters only differ in the number of dots with each other. This results in high sensitivity of any proposed OCR algorithm to noise.

- Some letter forms in Persian text have closed curves in their shapes, while others don't. This characteristic can be used as a feature to recognize some letters.

- There are different script styles for Persian language, e.g. Naskh, Nastaligh, Koofi, etc. The shapes of some letters are very different in these styles. Our focus in this paper is mainly on Naskh style, which is the dominant style in today's printed Persian text and in industrial applications.

We can approximate every Persian/Arabic letter using a set of primitive elements (or simply *primitives*) and dots [6, 7]. We can choose the set of primitives such that no two letters would be approximated with the same sequence of primitives. Thus, we would uniquely recognize any Persian/Arabic letter, given the sequence of its primitives and the number and the positions of its dots. Our proposed set of primitives consists of 8 elements, which are shown in Table II.

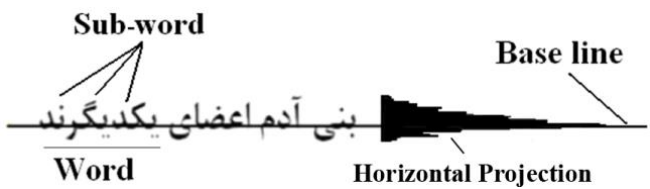

Figure 1. A Persian sentence showing a word, sub-words, horizontal projection of the sentence and baseline.

TABLE I. COMMON LETTER FORMS AND SIGNS IN FORMAL PERSIAN TEXT.

| Letter Name | Separate | First | Middle | End |
|---|---|---|---|---|
| Alef | ا | | | ا |
| Be | ب | بـ | ـبـ | ـب |
| Pe | پ | پـ | ـپـ | ـپ |
| Te | ت | تـ | ـتـ | ـت |
| Se | ث | ثـ | ـثـ | ـث |
| Jim | ج | جـ | ـجـ | ـج |
| Che | چ | چـ | ـچـ | ـچ |
| He (Jimi) | ح | حـ | ـحـ | ـح |
| Khe | خ | خـ | ـخـ | ـخ |
| Daal | د | | | ـد |
| Zaal | ذ | | | ـذ |
| Re | ر | | | ـر |
| Ze | ز | | | ـز |
| Zhe | ژ | | | ـژ |
| Sin | س | سـ | ـسـ | ـس |
| Shin | ش | شـ | ـشـ | ـش |
| Sad | ص | صـ | ـصـ | ـص |
| Zad | ض | ضـ | ـضـ | ـض |
| Ta | ط | ط | ـطـ | ـط |
| Za | ظ | ظ | ـظـ | ـظ |
| Eyn | ع | عـ | ـعـ | ـع |
| Gheyn | غ | غـ | ـغـ | ـغ |
| Fe | ف | فـ | ـفـ | ـف |
| Ghaaf | ق | قـ | ـقـ | ـق |
| Kaaf | ک | کـ | ـکـ | ـک |
| Gaaf | گ | گـ | ـگـ | ـگ |
| Laam | ل | لـ | ـلـ | ـل |
| Mim | م | مـ | ـمـ | ـم |
| Noon | ن | نـ | ـنـ | ـن |
| Vav | و | | | ـو |
| He (Docheshm) | ه | هـ | ـهـ | ـه |
| Ye | ی | یـ | ـیـ | ـی |
| Arabic Ye | ي | | | ـي |
| Fatheh sign | َ | | | |
| Kasreh sign | ِ | | | |
| Zammeh sign | ُ | | | |
| Tanvin Fatheh | ً | | | |
| Tanvin Kasreh | ٍ | | | |
| Tanvin Zammeh | ٌ | | | |
| Tashdid sign | ّ | | | |
| Mad sign | ~ | | | |
| Hamzeh sign | ء | | | |
| Hamzeh letter | ئ | ئـ | ـئـ | ـئ |
| *Total* | *43* | *26* | *26* | *34* |

TABLE II. THE SET OF PRIMITIVES IN PROPOSED OCR SYSTEM.

| Primitive Shape | Primitive Name | Primitive Code |
|---|---|---|
| ╲ | Backslash | B |
| ╱ | Slash | S |
| │ | Vertical | V |
| ─ | Horizontal | H |
| ⊂ | C Shape | C |
| ∩ | U Shape | U |
| ○ | Circle | O |
| ⌐ | Corner | L |

### III. PROPOSED OCR SYSTEM

The proposed OCR system has three main phases: preprocessing, main process and recognition. In this section, we will shortly describe the main steps of our OCR system. Fig. 2 shows the three main phases along with their sub-phases.

#### A. Preprocessing

The first step in almost every vision system is preprocessing. In this phase, the system makes the input image ready for the feature extraction. Each application requires its own type of preprocessing, e.g. preprocessing in industrial applications is different from preprocessing in banking applications. We assume that the input of the proposed OCR system is a scanned document, which is aligned and has mostly salt and pepper noise. Preprocessing in the OCR system has the following standard steps:

- *Binarization*: In this step, the input image is converted into a binary image. Our system uses a morphology-based non-uniform background removal method. This method works better than global thresholding on input images with illumination changes [8].

- *Filtering*: In this step, the binarized image is filtered using median filter to remove salt and pepper noise.

- *Normalization*: In this step, we estimate the font size by averaging the sizes of vertically connected black pixels of the filtered image and then use the estimated size to scale the input image. The aim of this step is to normalize the size of the text. This normalization makes the system more robust to font size changes.

Fig. 3 shows the result of preprocessing (i.e. binarization, filtering and normalization) on a sample bad-quality document.

#### B. Main Process

The aim of the second phase is to extract features of each sub-word in the normalized image. The features are primitives, and the number and positions of dots. The steps of the main process are:

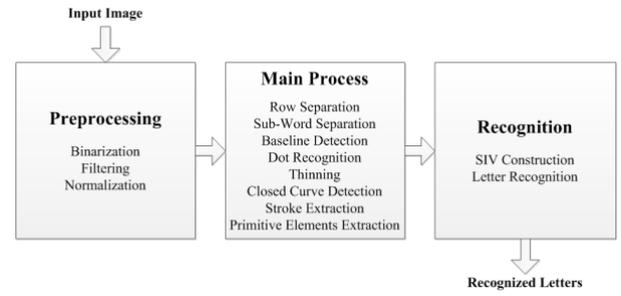

Figure 2. Phases of the proposed OCR system.

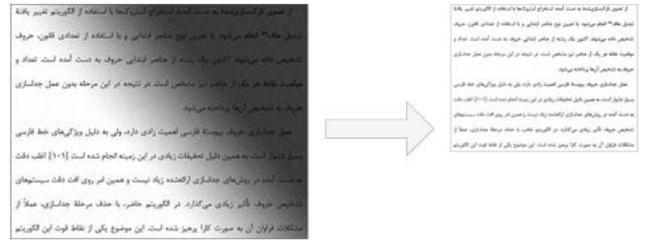

Figure 3. The result of preprocessing on a sample input document.

- *Row separation*: There is a space between consecutive rows. Since we assumed that the input document is aligned, we can use these spaces to separate text rows. Thus, the horizontal projection of the normalized image is used to separate text rows.

- *Sub-word separation*: Sub-words are separated using connected components algorithm. After extraction of all connected components in a line of text, we compare the sizes of all components with an empirical threshold. If the size of the component is above the threshold, we consider it a sub-word, otherwise it belongs to the vertically overlapping (or the nearest) sub-word.

- *Baseline detection*: In the horizontal projection of a typical text row, the image row with maximum value is the baseline of the text row. Thus, the horizontal projection of each separated row is used to detect the baseline of the row in our algorithm (as shown in Fig. 1).

- *Dot recognition*: In this step, the number and positions of dots are recognized. At first, we sort all the connected components of each sub-word by their area. The largest component is the main body of the sub-word. Other components are assumed to be dots and their number and positions is saved. The number of dots in each connected component is estimated based on the area and aspect ratio of the component.

- *Thinning*: We use the thinning algorithm of [9] to thin the main body of each sub-word, which is extracted from the dot recognition step. This thinning algorithm preserves connectivity, has unit-width convergence and approximates medial axis. Fig. 4 shows the result of this thinning algorithm on a Persian sentence.

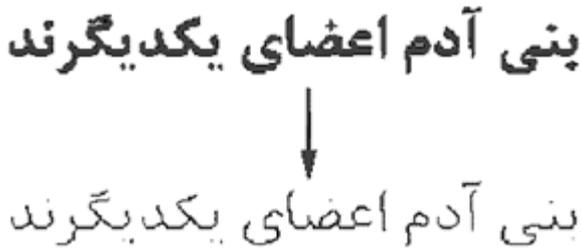

Figure 4. The result of thinning algorithm on a Persian sentence.

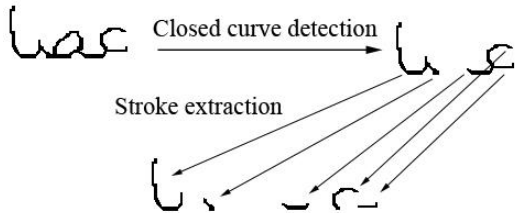

Figure 5. Extraction of closed curves and simple strokes from a sub-word.

- *Closed curve detection*: In this step, the number and position of closed curves (O primitive) in the thinned main body of a sub-word is detected using simple rules. We start following black pixels from the rightmost point until we reach an end-point or a branch point. If we reach an end-point, we conclude that this part does not have any closed curves. If we reach a branch point, we save the visited path and test every path starting from that branch point. If we reach this branch point or the visited path again, we know that there is a closed curve in the sub-word. Closed curves are deleted from the sub-word after this step. Fig. 5 shows the detection of closed curves in a sub-word.

- *Stroke extraction*: In this step, simple strokes are extracted from the thinned main body of each sub-word. A *stroke* is defined as a set of connected pixels in the thinned image. A *simple stroke* is a set of connected pixels between two branch (or end) points. For this step, we start following black pixels from each end-point and when we reach another end-point or branch point, we save the path as a simple stroke and remove it from the main body. Fig. 5 shows simple strokes extracted from a sub-word.

- *Primitive extraction*: Each stroke is approximated using a set of straight lines. For this step, the Modified Hough Transform is used [6]. We use 22.5° difference between line slopes; i.e. the angles of the lines are 0°, 22.5°, 45°, 67.5°, 90°, 112.5°, 135° and 157.5°. The extracted lines are then used to extract primitives using some simple rules. Fig. 6 shows the result of this step on a simple stroke.

Figure 6. Result of primitive extraction on a simple stroke.

*C. Recognition*

In the recognition phase, the extracted features of each sub-word are used to recognize the letters of the sub-word. The inputs of this phase are primitives and dots of each sub-word. This phase has two steps as follows:

- *Stroke Identification Vector (SIV) construction*: In this step, we construct a SIV for each stroke. Each SIV contains the primitives, number of dots above, number of dots below, horizontal start and end positions of the stroke and the vertical position of the stroke with respect to its baseline.

- *Letter Recognition*: Our system keeps one or more vectors for each Persian letter called Character Identification Vector (CIV). Each CIV contains the primitives used to approximate the letter, the number of dots above, the number of dots below and the position of the letter with respect to the baseline. An example CIV for the separate form of the letter "Shin" is shown in Fig. 7. In this step, SIVs from both ends of each sub-word are compared with CIVs until a letter is recognized. Then this letter is saved and the corresponding SIVs are removed. This procedure continues until all of the letters of the sub-word are recognized.

IV. TEST RESULTS

The proposed system is simulated using computer tools. Since there is no standard dataset for Persian OCR, we used the same dataset as we used in [7] to test the proposed OCR system. This dataset consists of 73818 sub-words. Table III shows the frequency of different lengths of sub-words in the dataset.

We tested the proposed OCR system on documents with 200 dpi resolution and 14 standard Persian fonts in 6 font sizes of 12, 14, 16, 18, 20 and 22. The fonts were Nazanin, Titr, Times New Roman, Mitra, Zar, Yaghut, Lotus, Badr, Compset, Jadid, Ferdosi, Roya, Traffic and Koodak.

The obtained precision was 97.06%. Table IV shows the precision of the system on different sub-word lengths.

| Letter Name | Letter | 1st Primitive | 2nd Primitive | 3rd Primitive | Points Above | Points Below | Position |
|---|---|---|---|---|---|---|---|
| Shin | ش | L | L | U | 3 | 0 | 0 |

Figure 7. CIV for separate form of letter "Shin".

TABLE III. THE DATASET USED TO TEST THE PROPOSED OCR SYSTEM.

| Number of letters in sub-word | Repetition in text | Repetition percent |
|---|---|---|
| 1 | 36657 | 49.66% |
| 2 | 23673 | 32.07% |
| 3 | 8952 | 12.13% |
| 4 | 3477 | 4.71% |
| 5 | 918 | 1.24% |
| 6 | 127 | 0.17% |
| 7 | 14 | 0.02% |
| Summation | 73818 | 100% |

TABLE IV. THE PRECISION OF THE PROPOSED OCR SYSTEM.

| Number of letters in sub-word | Repetition percent | Precision of the system |
|---|---|---|
| 1 | 49.66% | 99.8% |
| 2 | 32.07% | 96.2% |
| 3 | 12.13% | 93.3% |
| 4 | 4.71% | 87.9% |
| 5 or more | 1.43% | 83.0% |

The 97.06% precision of our method is much higher than 93.43% of the algorithm proposed in [7], which was tested on the same dataset. Since there is no standard dataset for testing Persian and Arabic OCR algorithms, we cannot fairly compare the results with other existing methods.

V. CONCLUSION

In this paper we introduced a model-based omnifont Persian/Arabic OCR system.

The proposed system assumed some properties for input documents. First, input document should be perfectly aligned; even a slight miss-alignment causes problems in row separation, which in turn results in the precision to drop significantly. Second assumption was in the normalization step: the system assumed that all characters in input document have the same size. Precision drops a little for documents with different font sizes. There were other assumptions, such as straight baseline, salt & pepper noise, etc. which are reasonable for a typical printed document. Most of these limitations can be eliminated by a better preprocessing phase.

The system used a set of 8 primitive elements as structural features for recognition. First, the scanned document was preprocessed. After normalizing the preprocessed image, text rows and sub-words were separated and then thinned. After recognition of dots in sub-words, strokes were extracted and primitive elements of each sub-word were recognized using the strokes. Finally, the primitives were compared with predefined set of character identification vectors in order to identify sub-word characters.

One of the advantages of the proposed system is that it does not need any prior knowledge about font types and sizes of the input document. This was achieved by using structural – instead of statistical – features (which are very sensitive to font changes). Also, the separation and recognition steps in the proposed OCR system are not totally separated, which makes the system less sensitive to native errors available in independent character separation process.

Another advantage of the introduced system is that it does not need training. This helps the system to be less sensitive to font changes when the fonts used in the training set are different from fonts used in the test set.

Test results showed that currently our introduced OCR system can be used for standard Persian fonts with a high accuracy. The system was not tested for Arabic text, but it is expected to work on Arabic text with a high accuracy as well. Although the system was not developed for recognition of handwritten letters, the omnifont nature of the system can be used to develop a robust Persian/Arabic handwriting recognition system.

REFERENCES

[1] E. Behjat, "Recognition of characters and shapes", MSc. Dissertation, Electrical Engineering Department, Sharif University of Technology, Tehran, Iran, 1975.

[2] M. Taraghi, "Recognition of printed persian characters using computer", MSc. Dissertation, Computer Science Department, Sharif University of Technology, Tehran, Iran, 1978.

[3] A. Adnan and A. Kaced, "Hand-written Arabic Character Recognition by the I.R.A.C. system", Proceedings of the 5th International Conference on Character Recognition, Miami Beach, December 1980, pp. 729-731.

[4] K. Badie and M. Shimura, "Machine recognition of Arabic cursive scripts", Pattern recognition in practice, Vol. 2, 1980, pp. 315-323.

[5] A. J. Alnsour and L. M. Alzoubady, "Arabic Handwritten Characters Recognized By Neocognitron Artificial Neural Network", University of Sharjah Journal of Pure & Applied Sciences, Vol. 3, June 2006, pp. 1-17.

[6] Sh. Ensafi, M. Miremadi, M. Eshghi, M. Naseri and A. Keipour, "Recognition of Separate and Adjoint Persian Letters in Less than Three Letter Subwords Using Primitives", Proceedings of Iran 17th Electrical Engineering Conference, Tehran, 11-13 May 2009.

[7] Sh. Ensafi, M. Eshghi and M. Naseri, "Recognition of separate and adjoint Persian letters using primitives", Proceedings of IEEE Symposium on Industrial Electronics & Applications, Vol. 2, Kuala Lumpur, 4-6 Oct 2009, pp. 611-616.

[8] P. Singh and A. K. Garg, "Non Uniform Background Removal using Morphology-based Structuring Element for Particle Analysis", International Journal of Computer Applications, Vol. 33, No. 6, pp. 11-16, Nov. 2011.

[9] B. K. Jang and R.T. Chin, "One-pass parallel thinning: analysis, properties, and quantitative evaluation", IEEE Transactions on Pattern Analysis and Machine Intelligence, Vol. 14, No. 11, Nov. 1992, pp. 1129-1140.